\RequirePackage[hyphens]{url}
\documentclass[11pt,a4paper, dvipsnames]{article}
\usepackage[hyperref]{emnlp2020}
\usepackage{times}
\usepackage{latexsym}

\usepackage{booktabs}
\usepackage{amsmath}
\usepackage{graphicx}
\usepackage{amssymb}
\usepackage{multirow}
\usepackage{xcolor}
\usepackage{tikz}
\usepackage{tabularx}

\usepackage{microtype}

\aclfinalcopy 

\definecolor{mygreen}{rgb}{0,.7,0}
\newcommand{\best}[1]{{\fontseries{b}\selectfont #1}}

\title{Inexpensive Domain Adaptation of Pretrained Language Models: \\ Case Studies on Biomedical NER and Covid-19 QA}
\author{Nina Poerner$^{\ast\dagger}$ and Ulli Waltinger$^\dagger$ and Hinrich Sch{\"u}tze$^\ast$ \\
  $^\ast$Center for Information and Language Processing, LMU Munich, Germany \\
  $^\dagger$Corporate Technology Machine Intelligence (MIC-DE), Siemens AG Munich, Germany \\
{\tt poerner@cis.uni-muenchen.de | inquiries@cislmu.org} }

\date{}

\newcounter{notecounter}
\newcommand{\enotesoff}{\long\gdef\enote##1##2{}}
\newcommand{\enoteson}{\long\gdef\enote##1##2{{
\stepcounter{notecounter}
{\large\bf
\hspace{1cm}\arabic{notecounter} $<<<$ ##1: ##2
$>>>$\hspace{1cm}}}}}
\enoteson
\enotesoff

\begin{document}
\maketitle
\begin{abstract}
\begingroup
\renewcommand{\footnotesize}{\tiny}
Domain adaptation of Pretrained Language Models (PTLMs) is
typically achieved by unsupervised pretraining on target-domain text.
While successful, this approach is expensive in terms of hardware, runtime and CO$_2$ emissions.
Here, we propose a cheaper alternative:
We train Word2Vec on target-domain text and align the resulting word vectors with the wordpiece vectors of a general-domain PTLM.
We evaluate on eight biomedical Named Entity Recognition (NER) tasks and compare against the recently proposed BioBERT model.
We cover over 60\% of the BioBERT -- BERT F1 delta, at 5\% of BioBERT's
CO$_2$ footprint and 2\% of its cloud compute cost.
We also show how to quickly adapt an existing general-domain Question Answering (QA) model to an emerging domain: the Covid-19 pandemic.\footnote{\url{www.github.com/npoe/covid-qa}}
\endgroup
\end{abstract}

\begin{table*}
\centering
\scriptsize
\setlength\tabcolsep{5pt}
\begin{tabularx}{.99\textwidth}{Xl|lrrrr}
\toprule
& size & Domain adaptation hardware &  Power(W) & Time(h) & CO$_2$(lbs) & Google Cloud \$ \\ \midrule
BioBERTv1.0 & base & 8 NVIDIA v100 GPUs (32GB) & 1505 & 240 & 544 & 1421 -- \phantom{0}4762\\
BioBERTv1.1 & base & 8 NVIDIA v100 GPUs (32GB) & 1505 & 552 & 1252 & 3268 -- 10952\\
GreenBioBERT (Section \ref{sec:experiment}) & base & 12 Intel Xeon E7-8857 CPUs, 30GB RAM & 1560 & 12 & 28 & \hfill 16 -- \phantom{000}76 \\
GreenCovidSQuADBERT (Section \ref{sec:experiment-qa}) & large & 12 Intel Xeon E7-8857 CPUs, 40GB RAM & 1560 & 24 & 56 & \hfill 32 -- \phantom{00}152 \\
\bottomrule
\end{tabularx}
\caption{Domain adaptation cost. CO$_2$ emissions are calculated according to \citet{strubell2019energy}. Since our hardware configuration is not available on Google Cloud, we take an \textit{m1-ultramem-40} instance (40 vCPUs, 961GB RAM) to estimate an upper bound on our Google Cloud cost.}
\label{tab:cost}
\end{table*}

\section{Introduction}
Pretrained Language Models (PTLMs) such as BERT \cite{devlin2019bert} have spearheaded advances on many NLP tasks.
Usually, PTLMs are pretrained on unlabeled general-domain and/or mixed-domain text, such as Wikipedia, digital books or the Common Crawl corpus.

When applying PTLMs to specific domains, it can be useful to domain-adapt them.
Domain adaptation of PTLMs has typically been achieved by
pretraining on target-domain text.
One such model is BioBERT \cite{lee2020biobert}, which was initialized from general-domain BERT and then pretrained on biomedical scientific publications.
The domain adaptation is shown to be helpful for target-domain tasks such as biomedical Named Entity Recognition (NER) or Question Answering (QA).
On the downside, the computational cost of pretraining can be considerable: 
BioBERTv1.0 was adapted for ten days on eight large GPUs (see Table \ref{tab:cost}), which is expensive, environmentally unfriendly, prohibitive for small research labs and students, and may delay prototyping on emerging domains.

We therefore propose a \textbf{fast, CPU-only domain-adaptation method for PTLMs}:
We train Word2Vec \cite{mikolov2013efficient} on target-domain text and align the resulting word vectors with the wordpiece vectors of an existing general-domain PTLM.
The PTLM thus gains domain-specific lexical knowledge in the form of additional word vectors, but its deeper layers remain unchanged.
Since Word2Vec and the vector space alignment are efficient models, the process requires a fraction of the resources associated with pretraining the PTLM itself, and it can be done on CPU.

In Section \ref{sec:experiment}, we use the proposed method to domain-adapt BERT on PubMed+PMC (the data used for BioBERTv1.0) and/or CORD-19 (Covid-19 Open Research Dataset).
We improve over general-domain BERT on eight out of eight biomedical NER tasks, using a fraction of the compute cost associated with BioBERT.
In Section \ref{sec:experiment-qa}, we show how to quickly adapt an existing Question Answering model to text about the Covid-19 pandemic, without any target-domain Language Model pretraining or finetuning.

\enote{hs}{below:  $\mathbb{L}$ not introduced?}

\section{Related work}
\subsection{The BERT PTLM}
\label{sec:ptlm}
For our purpose, a PTLM consists of three parts: 
A tokenizer $\mathcal{T}_\mathrm{LM} : \mathbb{L}^+ \rightarrow \mathbb{L}_\mathrm{LM}^+$, a wordpiece embedding lookup function $\mathcal{E}_\mathrm{LM}: \mathbb{L}_\mathrm{LM} \rightarrow \mathbb{R}^{d_\mathrm{LM}}$ and an encoder function $\mathcal{F}_\mathrm{LM}$.
$\mathbb{L}_\mathrm{LM}$ is a limited vocabulary of wordpieces.
All words from the natural language $\mathbb{L}^+$ that are not in $\mathbb{L}_\mathrm{LM}$ are tokenized into sequences of shorter wordpieces, e.g., \textit{dementia} becomes \textit{dem \#\#ent \#\#ia}.
Given a sentence $S = [w_1, \ldots, w_T]$, tokenized as $\mathcal{T}_\mathrm{LM}(S) = [\mathcal{T}_\mathrm{LM}(w_1); \ldots ; \mathcal{T}_\mathrm{LM}(w_T)]$, $\mathcal{E}_\mathrm{LM}$ embeds every wordpiece in $\mathcal{T}_\mathrm{LM}(S)$ into a real-valued, trainable wordpiece vector.
The wordpiece vectors of the entire sequence are stacked and fed into $\mathcal{F}_\mathrm{LM}$.
Note that we consider position and segment embeddings to be a part of $\mathcal{F}_\mathrm{LM}$ rather than $\mathcal{E}_\mathrm{LM}$.

In the case of BERT, $\mathcal{F}_\mathrm{LM}$ is a Transformer \cite{vaswani2017attention}, followed by a final Feed-Forward Net.
During pretraining, the Feed-Forward Net predicts the identity of masked wordpieces.
When finetuning on a supervised task, it is usually replaced with a randomly initialized layer.

\subsection{Domain-adapted PTLMs}
Domain adaptation of PTLMs is typically achieved by pretraining on unlabeled target-domain text.
Some examples of such models are BioBERT \cite{lee2020biobert}, which was pretrained on the PubMed and/or PubMed Central (PMC) corpora, Sci\-BERT \cite{beltagy2019scibert}, which was pretrained on papers from SemanticScholar, ClinicalBERT \cite{alsentzer2019publicly, huang2019clinicalbert} and ClinicalXLNet \cite{huang2019clinical}, which were pretrained on clinical patient notes, and AdaptaBERT \cite{han2019unsupervised}, which was pretrained on Early Modern English text.
In most cases, a domain-adapted PTLM is initialized from a general-domain PTLM (e.g., standard BERT), though \citet{beltagy2019scibert} report better results with a model that was pretrained from scratch with a custom wordpiece vocabulary.
In this paper, we focus on BioBERT, as its domain adaptation corpora are publicly available.

\subsection{Word vectors}
Word vectors are distributed representations of words that are trained on unlabeled text.
Contrary to PTLMs, word vectors are non-contextual, i.e., a word type is always assigned the same vector, regardless of context.
In this paper, we use Word2Vec \cite{mikolov2013efficient} to train word vectors.
We will denote the Word2Vec lookup function as $\mathcal{E}_\mathrm{W2V} : \mathbb{L}_\mathrm{W2V} \rightarrow \mathbb{R}^{d_\mathrm{W2V}}$.

\subsection{Word vector space alignment}
Word vector space alignment has most frequently been explored in the context of cross-lingual word embeddings.
For instance, \citet{mikolov2013exploiting} align English and Spanish Word2Vec spaces by a simple linear transformation.
\citet{wang2019improving} use a related method to align cross-lingual word vectors and multilingual BERT wordpiece vectors.

\section{Method}
\label{sec:method}
In the following, we assume access to a general-domain PTLM, as described in Section \ref{sec:ptlm}, and a corpus of unlabeled target-domain text.

\subsection{Creating new input vectors}
In a first step, we train Word2Vec on the target-domain corpus.
In a second step, we take the intersection of $\mathbb{L}_\mathrm{LM}$ and $\mathbb{L}_\mathrm{W2V}$. 
In practice, the intersection mostly contains wordpieces from $\mathbb{L}_\mathrm{LM}$ that correspond to standalone words.
It also contains single characters and other noise, however, we found that filtering them does not improve alignment quality.
In a third step, we use the intersection to fit an unconstrained linear transformation $\mathbf{W} \in \mathbb{R}^{d_\mathrm{LM} \times d_\mathrm{W2V}}$ via least squares:
$$
\underset{\mathbf{W}}{\mathrm{argmin}} \; \sum_{x \in \mathbb{L}_\mathrm{LM} \cap \mathbb{L}_\mathrm{W2V}} || \mathbf{W} \mathcal{E}_\mathrm{W2V}(x) - \mathcal{E}_\mathrm{LM}(x) ||_2^2
$$

\newcommand{\spc}{\hspace{.75mm}}

\definecolor{mygreen}{rgb}{0,.7,0}
\begin{table*}
\centering
\scriptsize
\setlength\tabcolsep{4pt}
\begin{tabularx}{.99\textwidth}{l|lXX}
\toprule
& \spc Query & 
NNs of query in \textcolor{blue}{$\mathcal{E}_\mathrm{LM}[\mathbb{L}_\mathrm{LM}]$} & 
NNs of query in \textcolor{mygreen}{$\mathbf{W} \mathcal{E}_\mathrm{W2V}[\mathbb{L}_\mathrm{W2V}]$} \\
\midrule
\multirow{4}{*}{\parbox{3.5cm}{query $\in \mathbb{L}_\mathrm{W2V} \cap \mathbb{L}_\mathrm{LM}$ \\ \\ \textbf{Boldface:} Training vector pairs}} & \spc \textcolor{blue}{\textbf{surgeon}} & 
\textcolor{blue}{physician, psychiatrist, surgery} & 
\textcolor{mygreen}{\textbf{surgeon}, urologist, neurosurgeon} \\

& \spc \textcolor{mygreen}{\textbf{surgeon}} & 
\textcolor{blue}{\textbf{surgeon}, physician, researcher} & 
\textcolor{mygreen}{neurosurgeon, urologist, radiologist} \\

& \spc \textcolor{blue}{\textbf{depression}} & 
\textcolor{blue}{Depression, recession, depressed} & 
\textcolor{mygreen}{\textbf{depression}, Depression, hopelessness} \\

& \spc \textcolor{mygreen}{\textbf{depression}} & 
\textcolor{blue}{\textbf{depression}, anxiety, anxiousness} & 
\textcolor{mygreen}{depressive, insomnia, Depression} \\ \midrule



\multirow{4}{*}{query $\in \mathbb{L}_\mathrm{W2V} - \mathbb{L}_\mathrm{LM}$} & 
\spc \textcolor{mygreen}{ventricular} & 
\textcolor{blue}{cardiac, pulmonary, mitochondrial} & 
\textcolor{mygreen}{atrial, ventricle, RV} \\


& \spc \textcolor{mygreen}{suppressants} & 
\textcolor{blue}{medications, medicines, medication} & 
\textcolor{mygreen}{suppressant, prokinetics, painkillers} \\

& \spc \textcolor{mygreen}{anesthesiologist} \hspace{5mm} & 
\textcolor{blue}{surgeon, technician, psychiatrist} & 
\textcolor{mygreen}{anesthetist, anaesthesiologist, anaesthetist} \\

& \spc \textcolor{mygreen}{nephrotoxicity} & 
\textcolor{blue}{toxicity, inflammation, contamination} & 
\textcolor{mygreen}{hepatotoxicity, ototoxicity, cardiotoxicity} \\



\bottomrule
\end{tabularx}
\begin{tabularx}{.99\textwidth}{Xc|ccc|c}
\toprule
& & BERT (ref) & BioBERTv1.0 (ref) & BioBERTv1.1 (ref) & GreenBioBERT \\
\multicolumn{2}{l|}{Biomedical NER task \hfill (NER task ID)} & 
\cite{lee2020biobert} & \cite{lee2020biobert} & \cite{lee2020biobert} & (with standard error of the mean)\\
\midrule
BC5CDR-disease \cite{li2016biocreative} & (1) & 81.97 / 82.48 / 82.41 & 85.86 / 87.27 / 86.56 & \best{86.47} / \best{87.84} / \best{87.15} 
& \underline{84.88} (.07) / \underline{85.29} (.12) / \underline{85.08} (.08) \\ 
NCBI-disease \cite{dougan2014ncbi} & (2) & 84.12 / \underline{87.19} / 85.63 & \best{89.04} / 89.69 / 89.36 & 88.22 / \best{91.25} / \best{89.71} 
& \underline{85.49} (.23) / 86.41 (.15) / \underline{85.94} (.16) \\ %

BC5CDR-chem \cite{li2016biocreative} & (3) & 90.94 / 91.38 / 91.16 & 93.27 / \best{93.61} / 93.44 & 93.68 / 93.26 / \best{93.47} 
& \underline{\best{93.82}} (.11) / \underline{92.35} (.17) / \underline{93.08} (.07)\\ %
BC4CHEMD \cite{krallinger2015chemdner} & (4) & 91.19 / 88.92 / 90.04 & 92.23 / 90.61 / 91.41 & \best{92.80} / \best{91.92} / \best{92.36} 
& \underline{\best{92.80}} (.04) / \underline{89.78} (.07) / \underline{91.26} (.04) \\
BC2GM \cite{smith2008overview} & (5) & 81.17 / 82.42 / 81.79 & \best{85.16} / 83.65 / 84.40 & 84.32 / \best{85.12} / \best{84.72} 
& \underline{83.34} (.15) / \underline{83.58} (.09) / \underline{83.45} (.10) \\
JNLPBA \cite{kim2004introduction} & (6) & 69.57 / 81.20 / 74.94 & \best{72.68} / 83.21 / \best{77.59} & 72.24 / \best{83.56} / 77.49 
& \underline{71.93} (.12) / \underline{82.58} (.12) / \underline{76.89} (.10) \\
LINNAEUS \cite{gerner2010linnaeus} & (7) & 91.17 / 84.30 / 87.60 & \best{93.84} / \best{86.11} / \best{89.81} & 90.77 / 85.83 / 88.24 
& \underline{92.50} (.17) / \underline{84.54} (.26) / \underline{88.34} (.18)\\
Species-800 \cite{pafilis2013species} & (8) & 69.35 / 74.05 / 71.63 & 72.84 / \best{77.97} / \best{75.31} & 72.80 / 75.36 / 74.06 
& \underline{\best{73.19}} (.26) / \underline{75.47} (.33) / \underline{74.31} (.24) \\
\bottomrule
\end{tabularx}
\caption{\best{Top:} Examples of within-space and cross-space nearest neighbors (NNs) by cosine similarity in GreenBioBERT's wordpiece embedding layer. \textcolor{blue}{Blue}: Original wordpiece space. \textcolor{mygreen}{Green}: Aligned Word2Vec space. \best{Bottom:} Biomedical NER test set precision / recall / F1 (\%). ``(ref)'': Reference scores from \citet{lee2020biobert}. \best{Boldface}: Best model in row. \underline{Underlined}: Best model without target-domain LM pretraining.}
\label{tab:results}
\end{table*}

\enote{hs}{probably hard to fix at this point, but hardly
  anyone will know these terms:
  \textcolor{mygreen}{VaD, MCI, AD}}

Intuitively, $\mathbf{W}$ makes Word2Vec vectors ``look like'' the PTLM's native wordpiece vectors, just like cross-lingual alignment makes word vectors from one language ``look like'' word vectors from another language.
In Table \ref{tab:results} (top), we show examples of within-space and cross-space nearest neighbors after alignment.

\subsection{Updating the wordpiece embedding layer}
\label{sec:inject}
Next, we redefine the wordpiece embedding layer of the PTLM.
The most radical strategy would be to replace the entire layer with the aligned Word2Vec vectors:
$$
\hat{\mathcal{E}}_\mathrm{LM} : \mathbb{L}_\mathrm{W2V} \rightarrow \mathbb{R}^{d_\mathrm{LM}} \; ; \;
\hat{\mathcal{E}}_\mathrm{LM}(x) = \mathbf{W} \mathcal{E}_\mathrm{W2V}(x)
$$

In initial experiments, this strategy led to a drop in
performance, presumably because
function words are not well represented by Word2Vec, and replacing them disrupts BERT's syntactic abilities.
To prevent this problem, we leave existing wordpiece vectors intact and only add new ones:
\begin{align}
\nonumber
\hat{\mathcal{E}}_\mathrm{LM} & : \mathbb{L}_\mathrm{LM} \cup \mathbb{L}_\mathrm{W2V} \rightarrow \mathbb{R}^{d_\mathrm{LM}}; \\
\label{eq:repl}
\hat{\mathcal{E}}_\mathrm{LM}(x) & = \begin{cases}
\mathcal{E}_\mathrm{LM}(x) & \text{ if } x \in \mathbb{L}_\mathrm{LM} \\
\mathbf{W} \mathcal{E}_\mathrm{W2V}(x) & \text{ otherwise}
\end{cases}
\end{align}

\subsection{Updating the tokenizer}
In a final step, we update the tokenizer to account for the added words.
Let $\mathcal{T}_\mathrm{LM}$ be the standard BERT tokenizer, and let $\hat{\mathcal{T}}_\mathrm{LM}$ be the tokenizer that treats all words in $\mathbb{L}_\mathrm{LM} \cup \mathbb{L}_\mathrm{W2V}$ as one-wordpiece tokens, while tokenizing any other words as usual.

In practice, a given
word may or may not benefit from being tokenized by
$\hat{\mathcal{T}}_\mathrm{LM}$ instead of $\mathcal{T}_\mathrm{LM}$.
To give a concrete example, 82\% of the words in the BC5CDR NER dataset that end in the suffix \textit{-ia} are 
part of
a disease entity (e.g., \textit{dementia}).
$\mathcal{T}_\mathrm{LM}$ tokenizes this word as \textit{dem \#\#ent \#\#ia}, thereby exposing this strong orthographic cue to the model.
As a result, $\mathcal{T}_\mathrm{LM}$ improves
recall on \textit{-ia} diseases.
But there are many cases where wordpiece tokenization is meaningless or misleading.
For instance \textit{euthymia} (not a disease) is tokenized by $\mathcal{T}_\mathrm{LM}$ as \textit{e \#\#uth \#\#ym \#\#ia}, making it likely to be classified as a disease.
By contrast, $\hat{\mathcal{T}}_\mathrm{LM}$ gives \textit{euthymia} a one-wordpiece representation that depends only on distributional semantics.
We find that using $\hat{\mathcal{T}}_\mathrm{LM}$ improves precision on \textit{-ia} diseases.

To combine these complementary strengths, we use a 50/50 mixture of $\mathcal{T}_\mathrm{LM}$-tokenization and $\hat{\mathcal{T}}_\mathrm{LM}$-tokenization when finetuning the PTLM on a task.
At test time, we use both tokenizers and mean-pool the outputs.
Let $o(\mathcal{T}(S))$ be some output of interest (e.g., a logit), given sentence $S$ tokenized by $\mathcal{T}$. We predict:
$\frac{1}{2} [o(\mathcal{T}_\mathrm{LM}(S)) + o(\hat{\mathcal{T}}_\mathrm{LM}(S))]$

\section{Experiment 1: Biomedical NER}
\label{sec:experiment}
In this section, we use the proposed method to create GreenBioBERT, an inexpensive and environmentally friendly alternative to BioBERT.
Recall that BioBERTv1.0 (\textit{biobert\_v1.0\_pubmed\_pmc}) was initialized from general-domain BERT (\textit{bert-base-cased}) and then pretrained on PubMed+PMC.

\subsection{Domain adaptation}
We train Word2Vec with vector size $d_\mathrm{W2V} = d_\mathrm{LM} = 768$ on PubMed+PMC (see Appendix for details).
Then, we update the wordpiece embedding layer and tokenizer of general-domain BERT (\textit{bert-base-cased}) as described in Section \ref{sec:method}.

\subsection{Finetuning}
We finetune GreenBioBERT on the eight publicly available NER tasks used in \citet{lee2020biobert}. 
We also do reproduction experiments with general-domain BERT and BioBERTv1.0, using the same setup as our model.
See Appendix for details on preprocessing and hyperparameters.
Since some of the datasets are sensitive to the random seed, we report mean and standard error over eight runs.

\subsection{Results and discussion}
\begin{figure}
\centering
\includegraphics[width=.99\columnwidth]{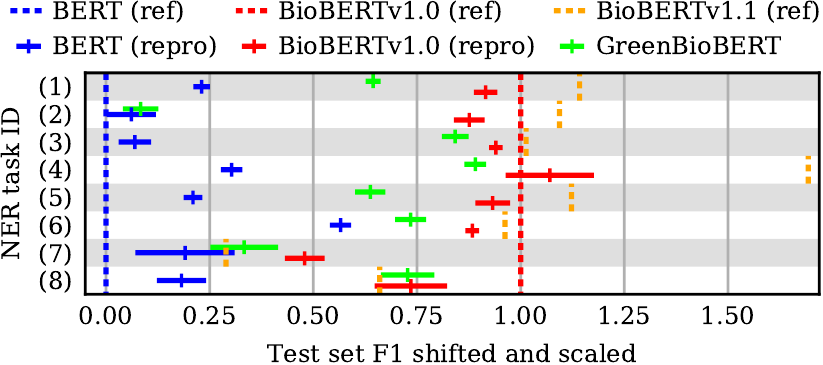}
\caption{NER test set F1, transformed as $(x - \mathrm{BERT}_\mathrm{(ref)}) / (\mathrm{BioBERTv1.0}_\mathrm{(ref)} - \mathrm{BERT}_\mathrm{(ref)})$. ``(ref)'': Reference scores from \citet{lee2020biobert}. ``(repro)'': Results of our reproduction experiments. Error bars: Standard error of the mean.}
\label{fig:scale}
\end{figure}

Table \ref{tab:results} (bottom) shows entity-level precision, recall and F1, as measured by the CoNLL NER scorer.
For ease of visualization, Figure \ref{fig:scale} shows what portion of the BioBERT -- BERT F1 delta is covered.
On average, we cover between 61\% and 70\% of the F1 delta (61\% for BioBERTv1.0 (ref), 70\% for BioBERTv1.1 (ref), and 61\% if we take our reproduction experiments as reference).

To test whether the improvements over general-domain BERT are due to the aligned Word2Vec vectors, or just to the availability of additional vectors in general, we perform an ablation study where we replace the aligned vectors with their non-aligned counterparts (by setting $\mathbf{W} = \mathbf{1}$ in Eq. \ref{eq:repl}) or with randomly initialized vectors. 
Table \ref{tab:qa-results} (top) shows that dev set F1 drops under these circumstances, i.e., vector space alignment is important.

\begingroup
\renewcommand{\footnotesize}{\tiny}
\section{Experiment 2: Covid-19 QA}
\label{sec:experiment-qa}
In this section, we use the proposed method to quickly adapt an existing general-domain QA model to an emerging target domain: Covid-19.
Our baseline model is SQuADBERT,\footnote{\url{www.huggingface.co/bert-large-uncased-whole-word-masking-finetuned-squad}} an existing BERT model that was finetuned on general-domain SQuAD \cite{rajpurkar2016squad}.
We evaluate on Deepset-AI Covid-QA \cite{moeller2020covid}, a SQuAD-style dataset with 2019 questions about 147 papers from CORD-19 (Covid-19 Open Research Dataset).
We assume that there is no labeled target-domain data for finetuning, which is a realistic setup for a new domain.
\endgroup

\subsection{Domain adaptation}
We train Word2Vec with vector size $d_\mathrm{W2V} = d_\mathrm{LM} = 1024$ on CORD-19 and/or PubMed+PMC.
The process takes less than an hour on CORD-19 and about one day on the combined corpus, again without the need for a GPU.
Then, we update SQuADBERT's wordpiece embedding layer and tokenizer, as described in Section \ref{sec:method}.
We refer to the resulting model as GreenCovidSQuADBERT.

\subsection{Results and discussion}
Table \ref{tab:qa-results} (bottom) shows that
GreenCovidSQuADBERT outperforms general-domain SQuADBERT on
all measures.
Interestingly, the small CORD-19 corpus is enough to achieve this result (compare ``CORD-19 only'' and ``CORD-19+PubMed+PMC''), presumably because it is specific to the target domain and contains the Covid-QA context papers.

\begin{table}
\centering
\scriptsize
\setlength{\tabcolsep}{3.25pt}
\begin{tabularx}{.99\columnwidth}{l|XXXXXXXX}
\toprule
NER task ID & \multicolumn{1}{c}{(1)} &  \multicolumn{1}{c}{(2)} &  \multicolumn{1}{c}{(3)} &  \multicolumn{1}{c}{(4)} &  \multicolumn{1}{c}{(5)} &  \multicolumn{1}{c}{(6)} &  \multicolumn{1}{c}{(7)} &  \multicolumn{1}{c}{(8)} \\\midrule
non-aligned & -4.88 & -3.50 & -4.13 & -3.34 & -2.34 & -0.56 & -0.84 & -4.63 \\
random init & -4.33 & -3.60 & -3.19 & -3.19 & -1.92 & -0.50 & -0.84 & -3.58 \\
\bottomrule
\end{tabularx}
\begin{tabularx}{.99\columnwidth}{l|X|ccc}
\toprule
& \multicolumn{1}{c|}{domain adaptation corpus\hfill size} & EM & F1 & substr \\ \midrule
SQuADBERT & \multicolumn{1}{c|}{--------} & 33.04 & 58.24 & 65.87 \\ \midrule
GreenCovid- & CORD-19 only \hfill 2GB &  \best{34.62} & 60.09 & \best{68.20} \\
SQuADBERT & CORD-19+PubMed+PMC \hfill 94GB & 34.32 & \best{60.23} & 68.00 \\
\bottomrule
\end{tabularx}
\caption{\best{Top:} NER ablation study. Drop in dev set F1 (w.r.t. GreenBioBERT) when using non-aligned or randomly initialized word vectors instead of aligned word vectors. \best{Bottom:} Results (\%) on Deepset-AI Covid-QA. EM (exact match) and F1 are evaluated with the SQuAD scorer. ``substr'': Predictions that are a substring of the gold answer. Much higher than EM, because many gold answers are not minimal answer spans (see Appendix, ``Notes on Covid-QA'', for an example).}
\label{tab:qa-results}
\end{table}

\section{Conclusion}
As a reaction to the trend towards high-resource models, we have proposed an inexpensive, CPU-only method for domain-adapting Pretrained Language Models:
We train Word2Vec vectors on target-domain data and align them with the wordpiece vector space of a general-domain PTLM.

On eight biomedical NER tasks, we cover over 60\% of the BioBERT -- BERT F1 delta, at 5\% of BioBERT's domain adaptation CO$_2$ footprint and 2\% of its cloud compute cost.
We have also shown how to rapidly adapt an existing BERT QA model to an emerging domain -- the Covid-19 pandemic -- without the need for target-domain Language Model pretraining or finetuning.

We hope that our approach will benefit practitioners with limited time or resources, and that it will encourage environmentally friendlier NLP.

\bibliography{emnlp2020}

\begin{thebibliography}{23}
\expandafter\ifx\csname natexlab\endcsname\relax\def\natexlab#1{#1}\fi

\bibitem[{Alsentzer et~al.(2019)Alsentzer, Murphy, Boag, Weng, Jindi, Naumann,
  and McDermott}]{alsentzer2019publicly}
Emily Alsentzer, John Murphy, William Boag, Wei-Hung Weng, Di~Jindi, Tristan
  Naumann, and Matthew McDermott. 2019.
\newblock \href {https://doi.org/10.18653/v1/w19-1909} {Publicly available
  clinical {BERT} embeddings}.
\newblock In \emph{2nd Clinical Natural Language Processing Workshop}, pages
  72--78, Minneapolis, USA.

\bibitem[{Beltagy et~al.(2019)Beltagy, Lo, and Cohan}]{beltagy2019scibert}
Iz~Beltagy, Kyle Lo, and Arman Cohan. 2019.
\newblock \href {https://doi.org/10.18653/v1/d19-1371} {{SciBERT}: A pretrained
  language model for scientific text}.
\newblock In \emph{EMNLP-IJCNLP}, pages 3606--3611, Hong Kong, China.

\bibitem[{Devlin et~al.(2019)Devlin, Chang, Lee, and
  Toutanova}]{devlin2019bert}
Jacob Devlin, Ming-Wei Chang, Kenton Lee, and Kristina Toutanova. 2019.
\newblock {BERT}: Pre-training of deep bidirectional transformers for language
  understanding.
\newblock In \emph{NAACL-HLT}, pages 4171--4186, Minneapolis, USA.

\bibitem[{Dodge et~al.(2019)Dodge, Gururangan, Card, Schwartz, and
  Smith}]{dodge2019show}
Jesse Dodge, Suchin Gururangan, Dallas Card, Roy Schwartz, and Noah~A Smith.
  2019.
\newblock \href {https://doi.org/10.18653/v1/d19-1224} {Show your work:
  Improved reporting of experimental results}.
\newblock In \emph{EMNLP-IJCNLP}, pages 2185--2194, Hong Kong, China.

\bibitem[{Do{\u{g}}an et~al.(2014)Do{\u{g}}an, Leaman, and Lu}]{dougan2014ncbi}
Rezarta~Islamaj Do{\u{g}}an, Robert Leaman, and Zhiyong Lu. 2014.
\newblock \href {https://doi.org/10.1016/j.jbi.2013.12.006} {{NCBI} disease
  corpus: a resource for disease name recognition and concept normalization}.
\newblock \emph{Journal of biomedical informatics}, 47:1--10.

\bibitem[{Gerner et~al.(2010)Gerner, Nenadic, and Bergman}]{gerner2010linnaeus}
Martin Gerner, Goran Nenadic, and Casey~M Bergman. 2010.
\newblock \href {https://doi.org/10.1186/1471-2105-11-85} {{LINNAEUS}: a
  species name identification system for biomedical literature}.
\newblock \emph{BMC bioinformatics}, 11(1):85.

\bibitem[{Han and Eisenstein(2019)}]{han2019unsupervised}
Xiaochuang Han and Jacob Eisenstein. 2019.
\newblock \href {https://doi.org/10.18653/v1/d19-1433} {Unsupervised domain
  adaptation of contextualized embeddings for sequence labeling}.
\newblock In \emph{EMNLP-IJCNLP}, pages 4229--4239, Hong Kong, China.

\bibitem[{Huang et~al.(2019{\natexlab{a}})Huang, Altosaar, and
  Ranganath}]{huang2019clinicalbert}
Kexin Huang, Jaan Altosaar, and Rajesh Ranganath. 2019{\natexlab{a}}.
\newblock \href {https://arxiv.org/abs/1904.05342} {Clinical{BERT}: Modeling
  clinical notes and predicting hospital readmission}.
\newblock \emph{arXiv preprint arXiv:1904.05342}.

\bibitem[{Huang et~al.(2019{\natexlab{b}})Huang, Singh, Chen, Moseley, Deng,
  George, and Lindvall}]{huang2019clinical}
Kexin Huang, Abhishek Singh, Sitong Chen, Edward~T Moseley, Chih-ying Deng,
  Naomi George, and Charlotta Lindvall. 2019{\natexlab{b}}.
\newblock \href {https://arxiv.org/pdf/1912.11975} {Clinical {XLNet}: Modeling
  sequential clinical notes and predicting prolonged mechanical ventilation}.
\newblock \emph{arXiv preprint arXiv:1912.11975}.

\bibitem[{Kim et~al.(2004)Kim, Ohta, Tsuruoka, Tateisi, and
  Collier}]{kim2004introduction}
Jin-Dong Kim, Tomoko Ohta, Yoshimasa Tsuruoka, Yuka Tateisi, and Nigel Collier.
  2004.
\newblock \href {https://doi.org/10.3115/1567594.1567610} {Introduction to the
  bio-entity recognition task at {JNLPBA}}.
\newblock In \emph{International Joint Workshop on Natural Language Processing
  in Biomedicine and its Applications}, pages 70--75.

\bibitem[{Krallinger et~al.(2015)Krallinger, Rabal, Leitner, Vazquez, Salgado,
  Lu, Leaman, Lu, Ji, Lowe et~al.}]{krallinger2015chemdner}
Martin Krallinger, Obdulia Rabal, Florian Leitner, Miguel Vazquez, David
  Salgado, Zhiyong Lu, Robert Leaman, Yanan Lu, Donghong Ji, Daniel~M Lowe,
  et~al. 2015.
\newblock The {CHEMDNER} corpus of chemicals and drugs and its annotation
  principles.
\newblock \emph{Journal of cheminformatics}, 7(1):1--17.

\bibitem[{Lee et~al.(2020)Lee, Yoon, Kim, Kim, Kim, So, and
  Kang}]{lee2020biobert}
Jinhyuk Lee, Wonjin Yoon, Sungdong Kim, Donghyeon Kim, Sunkyu Kim, Chan~Ho So,
  and Jaewoo Kang. 2020.
\newblock \href {https://doi.org/10.1093/bioinformatics/btz682} {{BioBERT}: A
  pre-trained biomedical language representation model for biomedical text
  mining}.
\newblock \emph{Bioinformatics}, 36(4):1234--1240.

\bibitem[{Li et~al.(2016)Li, Sun, Johnson, Sciaky, Wei, Leaman, Davis,
  Mattingly, Wiegers, and Lu}]{li2016biocreative}
Jiao Li, Yueping Sun, Robin~J Johnson, Daniela Sciaky, Chih-Hsuan Wei, Robert
  Leaman, Allan~Peter Davis, Carolyn~J Mattingly, Thomas~C Wiegers, and Zhiyong
  Lu. 2016.
\newblock \href {https://doi.org/10.1093/database/baw068} {{BioCreative V CDR}
  task corpus: a resource for chemical disease relation extraction}.
\newblock \emph{Database}, 2016.

\bibitem[{Loshchilov and Hutter(2018)}]{loshchilov2018fixing}
Ilya Loshchilov and Frank Hutter. 2018.
\newblock Fixing weight decay regularization in {Adam}.

\bibitem[{Mikolov et~al.(2013{\natexlab{a}})Mikolov, Chen, Corrado, and
  Dean}]{mikolov2013efficient}
Tomas Mikolov, Kai Chen, Greg Corrado, and Jeffrey Dean. 2013{\natexlab{a}}.
\newblock \href {https://arxiv.org/abs/1301.3781} {Efficient estimation of word
  representations in vector space}.
\newblock \emph{arXiv preprint arXiv:1301.3781}.

\bibitem[{Mikolov et~al.(2013{\natexlab{b}})Mikolov, Le, and
  Sutskever}]{mikolov2013exploiting}
Tomas Mikolov, Quoc~V Le, and Ilya Sutskever. 2013{\natexlab{b}}.
\newblock \href {https://arxiv.org/abs/1309.4168} {Exploiting similarities
  among languages for machine translation}.
\newblock \emph{arXiv preprint arXiv:1309.4168}.

\bibitem[{M{\"o}ller et~al.(2020)M{\"o}ller, Reina, Jayakumar, and
  Pietsch}]{moeller2020covid}
Timo M{\"o}ller, Anthony Reina, Raghavan Jayakumar, and Malte Pietsch. 2020.
\newblock Covid-qa: A question \& answer dataset for covid-19.

\bibitem[{Pafilis et~al.(2013)Pafilis, Frankild, Fanini, Faulwetter, Pavloudi,
  Vasileiadou, Arvanitidis, and Jensen}]{pafilis2013species}
Evangelos Pafilis, Sune~P Frankild, Lucia Fanini, Sarah Faulwetter, Christina
  Pavloudi, Aikaterini Vasileiadou, Christos Arvanitidis, and Lars~Juhl Jensen.
  2013.
\newblock \href {https://doi.org/10.1371/journal.pone.0065390} {The {SPECIES}
  and {ORGANISMS} resources for fast and accurate identification of taxonomic
  names in text}.
\newblock \emph{{PloS} one}, 8(6).

\bibitem[{Rajpurkar et~al.(2016)Rajpurkar, Zhang, Lopyrev, and
  Liang}]{rajpurkar2016squad}
Pranav Rajpurkar, Jian Zhang, Konstantin Lopyrev, and Percy Liang. 2016.
\newblock \href {https://doi.org/10.18653/v1/d16-1264} {{SQuAD}: 100,000+
  questions for machine comprehension of text}.
\newblock In \emph{EMNLP}, pages 2383--2392, Austin, USA.

\bibitem[{Smith et~al.(2008)Smith, Tanabe, nee Ando, Kuo, Chung, Hsu, Lin,
  Klinger, Friedrich, Ganchev et~al.}]{smith2008overview}
Larry Smith, Lorraine~K Tanabe, Rie~Johnson nee Ando, Cheng-Ju Kuo, I-Fang
  Chung, Chun-Nan Hsu, Yu-Shi Lin, Roman Klinger, Christoph~M Friedrich, Kuzman
  Ganchev, et~al. 2008.
\newblock \href {https://doi.org/10.1186/gb-2008-9-s2-s2} {Overview of
  {BioCreative II} gene mention recognition}.
\newblock \emph{Genome biology}, 9(2):S2.

\bibitem[{Strubell et~al.(2019)Strubell, Ganesh, and
  McCallum}]{strubell2019energy}
Emma Strubell, Ananya Ganesh, and Andrew McCallum. 2019.
\newblock \href {https://doi.org/10.18653/v1/p19-1355} {Energy and policy
  considerations for deep learning in {NLP}}.
\newblock In \emph{ACL}, pages 3645--3650, Florence, Italy.

\bibitem[{Vaswani et~al.(2017)Vaswani, Shazeer, Parmar, Uszkoreit, Jones,
  Gomez, Kaiser, and Polosukhin}]{vaswani2017attention}
Ashish Vaswani, Noam Shazeer, Niki Parmar, Jakob Uszkoreit, Llion Jones,
  Aidan~N Gomez, {\L}ukasz Kaiser, and Illia Polosukhin. 2017.
\newblock Attention is all you need.
\newblock In \emph{NeurIPS}, pages 5998--6008, Long Beach, USA.

\bibitem[{Wang et~al.(2019)Wang, Yu, Sun, Chen, and Yu}]{wang2019improving}
Hai Wang, Dian Yu, Kai Sun, Janshu Chen, and Dong Yu. 2019.
\newblock \href {https://doi.org/10.18653/v1/K19-1030} {Improving pre-trained
  multilingual models with vocabulary expansion}.
\newblock In \emph{CoNLL}, pages 316--327, Hong Kong, China.

\end{thebibliography}
\bibliographystyle{acl_natbib}

\clearpage

\appendix

\section*{Inexpensive Domain Adaptation of Pretrained Language Models (Appendix)}
\subsection*{Word2Vec training}
We downloaded the PubMed, PMC and CORD-19 corpora from:
\begin{itemize}
\small
\setlength\itemsep{0mm}
\item \url{https://ftp.ncbi.nlm.nih.gov/pub/pmc/oa_bulk/} [20 January 2020, 68GB raw text]
\item \url{https://ftp.ncbi.nlm.nih.gov/pubmed/baseline/} [20 January 2020, 24GB raw text]
\item \url{https://pages.semanticscholar.org/coronavirus-research} [17 April 2020, 2GB raw text]
\end{itemize}
We extract all abstracts and text bodies and apply the BERT basic tokenizer (a rule-based word tokenizer that standard BERT uses before wordpiece tokenization).
Then, we train CBOW Word2Vec\footnote{\url{www.github.com/tmikolov/word2vec}} with negative sampling.
We use default parameters except for the vector size (which we set to $d_\mathrm{W2V} = d_\mathrm{LM}$).

\subsection*{Experiment 1: Biomedical NER}
\subsubsection*{Pretrained models}
General-domain BERT and BioBERTv1.0 were downloaded from:
\begin{itemize}
\small
\setlength\itemsep{0mm}
\item \url{www.storage.googleapis.com/bert_models/2018_10_18/cased_L-12_H-768_A-12.zip}
\item \url{www.github.com/naver/biobert-pretrained}
\end{itemize}

\subsubsection*{Data}
We downloaded the NER datasets by following instructions on \url{www.github.com/dmis-lab/biobert#Datasets}.
For detailed dataset statistics, see \citet{lee2020biobert}.

\subsubsection*{Preprocessing}
We use \citet{lee2020biobert}'s preprocessing strategy:
We cut all sentences into chunks of 30 or fewer whitespace-tokenized words (without splitting inside labeled spans).
Then, we tokenize every chunk $S$ with $\mathcal{T} = \mathcal{T}_\mathrm{LM}$ or $\mathcal{T} = \hat{\mathcal{T}}_\mathrm{LM}$ and add special tokens:
$$ X = \textit{[CLS]} \; \mathcal{T}(S) \; \textit{[SEP]} $$
Word-initial wordpieces in $\mathcal{T}(S)$ are labeled as \textit{B(egin)}, \textit{I(nside)} or \textit{O(utside)}, while non-word-initial wordpieces are labeled as \textit{X(ignore)}.

\subsubsection*{Modeling, training and inference}
We follow \citet{lee2020biobert}'s implementation (\url{www.github.com/dmis-lab/biobert}):
We add a randomly initialized softmax classifier on top of the last BERT layer to predict the labels.
We finetune the entire model to minimize negative log likelihood, with the AdamW optimizer \cite{loshchilov2018fixing} and a linear learning rate scheduler (10\% warmup).
All finetuning runs were done on a GeForce Titan X GPU (12GB).

At inference time, we gather the output logits of word-initial wordpieces only.
Since the number of word-initial wordpieces is the same for $\mathcal{T}_\mathrm{LM}(S)$ and $\hat{\mathcal{T}}_\mathrm{LM}(S)$, this makes mean-pooling the logits straightforward.

\subsubsection*{Hyperparameters}
\begin{figure*}
\centering
\includegraphics[width=.99\textwidth, trim=10mm 1mm 1mm 1mm, clip]{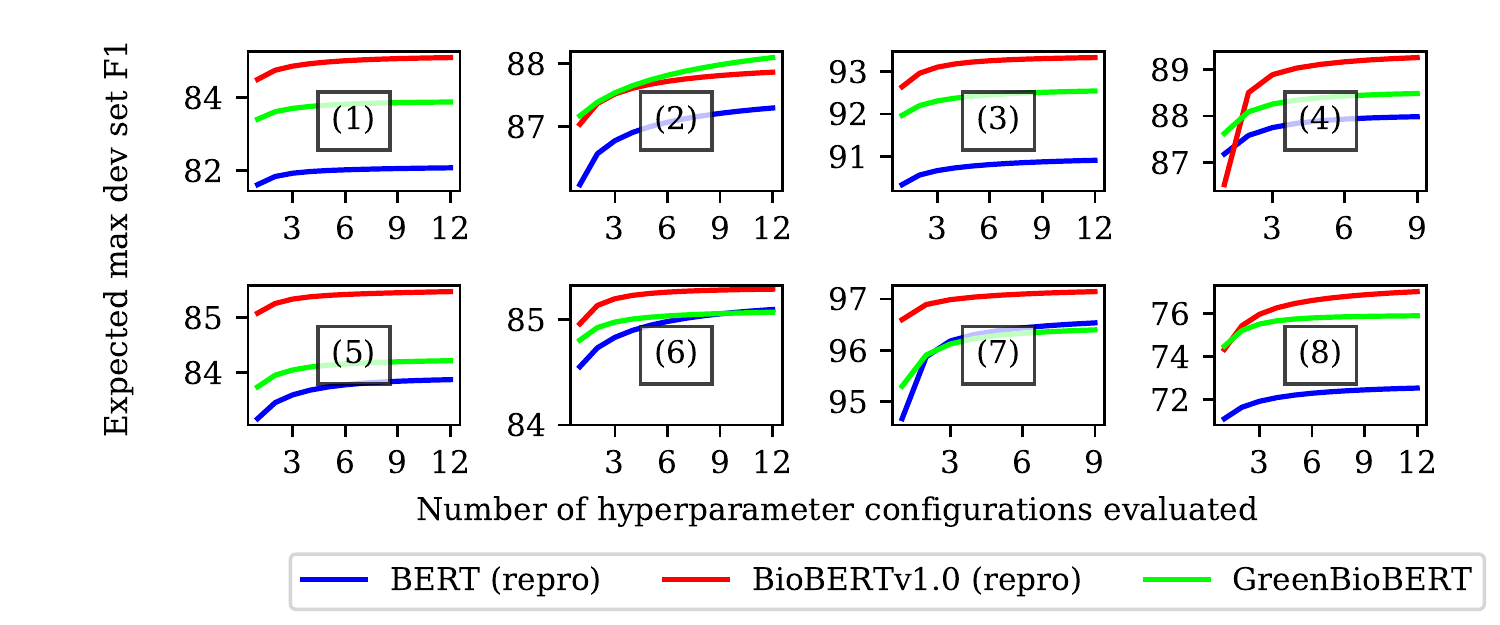}
\caption{Expected maximum F1 on NER development sets as a function of the number of evaluated hyperparameter configurations. Numbers in brackets are NER task IDs (see Table \ref{tab:hyper}).}
\label{fig:dodge}
\end{figure*}

We tune the batch size and peak learning rate on the development set (metric: F1), using the same hyperparameter space as \citet{lee2020biobert}:
\begin{description}
	\item[Batch size:] $[10, 16, 32, 64]$\footnote{Since LINNAEUS and BC4CHEM have longer maximum tokenized chunk lengths than the other datasets, our hardware was insufficient to evaluate batch size 64 on them.}
	\item[Learning rate:] $[1\cdot10^{-5}, 3\cdot10^{-5}, 5\cdot10^{-5}]$
\end{description}
We train for $100$ epochs, which is the upper end of the 50--100 range recommended by the original authors.
After selecting the best configuration for every task and model (see Table \ref{tab:hyper}), we train the final model on the concatenation of training and development set, as was done by \citet{lee2020biobert}.
See Figure \ref{fig:dodge} for expected maximum development set F1 as a function of the number of evaluated hyperparameter configurations \cite{dodge2019show}.

\subsection*{Experiment 2: Covid-19 QA}
\subsubsection*{Pretrained model}
We downloaded the SQuADBERT baseline from:
\begin{itemize}
\small
\setlength\itemsep{0mm}
\item \url{www.huggingface.co/bert-large-uncased-whole-word-masking-finetuned-squad}
\end{itemize}

\subsubsection*{Data}
We downloaded the Deepset-AI Covid-QA dataset from:
\begin{itemize}
\small
\setlength\itemsep{0mm}
\item \url{www.github.com/deepset-ai/COVID-QA/blob/master/data/question-answering/COVID-QA.json} [24 June 2020]
\end{itemize}

At the time of writing, the dataset contains 2019 questions and gold answer spans.\footnote{In an earlier version of the paper, we reported results on a preliminary version of Deepset-AI Covid-QA, which contained 1380 questions.}
Every question is associated with one of 147 research papers (contexts) from CORD-19.\footnote{\url{www.github.com/deepset-ai/COVID-QA/issues/103}}
Since we do not do target-domain finetuning, we treat the entire dataset as a test set.

\subsubsection*{Preprocessing}
We tokenize every question-context pair $(Q, C)$ with $\mathcal{T} = \mathcal{T}_\mathrm{LM}$ or $\mathcal{T} = \hat{\mathcal{T}}_\mathrm{LM}$, which yields $(\mathcal{T}(Q), \mathcal{T}(C))$.
Since $\mathcal{T}(C)$ is usually too long to be digested in a single forward pass, we define a sliding window with width and stride $N = \mathrm{floor}(\frac{509 - |\mathcal{T}(Q)|}{2})$.
At step $n$, the ``active'' window is between $a^{(l)}_n = (n-1)N+1$ and $a^{(r)}_n = \mathrm{min}(|C|, nN)$.
The input is defined as:
\begin{align} \nonumber
X^{(n)} = & \; \textit{[CLS]} \; \mathcal{T}(Q) \; \textit{[SEP]} \\ \nonumber
& \; \mathcal{T}(C)_{a^{(l)}_n-p^{(l)}_n:a^{(r)}_n+p^{(r)}_n} \; \textit{[SEP]}
\end{align}
$p^{(l)}_n$ and $p^{(r)}_n$ are chosen such that $|X^{(n)}| = 512$, and such that the active window is in the center of the input (if possible).

\subsubsection*{Modeling and inference}
Feeding $X^{(n)}$ into the QA model yields start logits $\mathbf{h'}^{(\mathrm{start}, n)} \in \mathbb{R}^{|X^{(n)}|}$ and end logits $\mathbf{h'}^{(\mathrm{end},n)} \in \mathbb{R}^{|X^{(n)}|}$.
We extract and concatenate the slices that correspond to the active windows of all steps:
\begin{align} \nonumber
\mathbf{h}^{(*)} & \in  \mathbb{R}^{|\mathcal{T}(C)|} \\ \nonumber
\mathbf{h}^{(*)} & = [\mathbf{h}'^{(*, 1)}_{a^{(l)}_1:a^{(r)}_1}; \ldots; \mathbf{h}'^{(*, n)}_{a^{(l)}_n:a^{(r)}_n}; \ldots]
\end{align}

Next, we map the logits from the wordpiece level to the word level.
This allows us to mean-pool the outputs of $\mathcal{T}_\mathrm{LM}$ and $\hat{\mathcal{T}}_\mathrm{LM}$ even when $|\mathcal{T}_\mathrm{LM}(C)| \neq |\hat{\mathcal{T}}_\mathrm{LM}(C)|$.

Let $c_i$ be a word in $C$ and let $\mathcal{T}(C)_{j:j+|\mathcal{T}(c_i)|}$ be the corresponding wordpieces.
The start and end logits of $c_i$ are:
$$o^{(*)}_i = \mathrm{max}_{j \leq j' \leq j+|\mathcal{T}(c_i)|} [h^{(*)}_{j'}]$$

Finally, we return the answer span $C_{k:k'}$ that maximizes $o^{(\mathrm{start})}_k + o^{(\mathrm{end})}_{k'}$, subject to the constraints that $k'$ does not precede $k$ and the answer contains no more than 500 characters.

\subsubsection*{Notes on Covid-QA}
There are some important differences between Covid-QA and SQuAD, which make the task challenging:
\begin{itemize}
	\item The Covid-QA contexts are full documents rather than single paragraphs. Thus, the correct answer may appear several times, often with slightly different wordings.
	But only a single occurrence is annotated as correct, e.g.:
		\begin{description}
			\item[Question:] What was the prevalence of Coronavirus OC43 in community samples in Ilorin, Nigeria?
			\item[Correct:] 13.3\% (95\% CI 6.9-23.6\%) \textit{\# from main text}
			\item[Predicted:] 13.3\%, 10/75 \textit{\# from abstract}
		\end{description}		
	\item SQuAD gold answers are defined as the ``shortest span in the paragraph that answered the question'' \cite[p. 4]{rajpurkar2016squad}, but many Covid-QA gold answers are longer and contain non-essential context, e.g.:
	\begin{description}
		\item[Question:] When was the  Middle East Respiratory Syndrome Coronavirus isolated first?
		\item[Correct:] (MERS-CoV) was first isolated in 2012, in a 60-year-old man who died in Jeddah, KSA due to severe acute pneumonia and multiple organ failure
		\item[Predicted:] 2012
	\end{description}
\end{itemize}
These differences are part of the reason why the exact match score is lower than the word-level F1 score and the substring score (see Table \ref{tab:qa-results}, bottom, main paper).

\begin{table*}
\centering
\small
\begin{tabularx}{.99\textwidth}{Xr|cc|cc|cc}
	\toprule
	&& \multicolumn{2}{c|}{BERT (repro)} & \multicolumn{2}{c|}{BioBERTv1.0 (repro)} & \multicolumn{2}{c}{GreenBioBERT} \\
	Biomedical NER task & (ID) & hyperparams & dev set F1 & hyperparams & dev set F1 & hyperparams & dev set F1 \\
	\midrule
	BC5CDR-disease & (1) & $32, 3\cdot10^{-5}$ & $82.12$ & $10, 1\cdot10^{-5}$ & $85.15$ & $32, 1\cdot10^{-5}$ & $83.90$ \\
	NCBI-disease & (2) & $32, 3\cdot10^{-5}$ & $87.52$ & $32, 1\cdot10^{-5}$ & $87.99$ & $10, 3\cdot10^{-5}$ & $88.43$ \\
	BC5CDR-chem & (3) & $64, 3\cdot10^{-5}$ & $91.00 $ & $32, 1\cdot10^{-5}$ & $93.36$ & $10, 1\cdot10^{-5}$ & $92.59$ \\
	BC4CHEMD & (4) & $16, 1\cdot10^{-5}$ & $88.02$ & $32, 1\cdot10^{-5}$ & $89.35$ & $16, 1\cdot10^{-5}$ & $88.53$ \\
	BC2GM & (5) & $32, 1\cdot10^{-5}$ & $83.91$ & $64, 3\cdot10^{-5}$ & $85.54$ & $64, 3\cdot10^{-5}$ & $84.25$ \\
	JNLPBA & (6) & $32, 5\cdot10^{-5}$ & $85.18$ & $32, 5\cdot10^{-5}$ & $85.30$ & $10, 3\cdot10^{-5}$ & $ 85.10$ \\
	LINNAEUS & (7) & $16, 1\cdot10^{-5}$ & $96.67$ & $32, 1\cdot10^{-5}$ & $97.22$ & $10, 1\cdot10^{-5}$ & $96.49$ \\
	Species-800 & (8) & $32, 1\cdot10^{-5}$ & $72.70$ & $32, 1\cdot10^{-5}$ & $77.34$ & $16, 1\cdot10^{-5}$ & $75.93$ \\
	\bottomrule
\end{tabularx}
\caption{Best hyperparameters (batch size, peak learning rate) and best dev set F1 per NER task and model. BERT (repro) and BioBERTv1.0 (repro) refer to our reproduction experiments.}
\label{tab:hyper}
\end{table*}

\end{document}